УДК 004.415

**Палагин А.В., Петренко Н.Г., Величко В.Ю., Малахов К.С.**

**К ВОПРОСУ РАЗРАБОТКИ ОНТОЛОГО-УПРАВЛЯЕМОЙ АРХИТЕКТУРЫ ИНТЕЛЛЕКТУАЛЬНОЙ ПРОГРАММНОЙ СИСТЕМЫ**

В работе описана архитектура интеллектуальной программной системы автоматизированного построения онтологических баз знаний предметных областей и программная модель подсистемы управляющей графической оболочки.

*Ключевые слова:* онтология предметной области, инструментальный комплекс онтологического назначения, управляющая графическая оболочка.

**Введение**

Одной из основных проблем при создании систем обработки знаний является проблема автоматизированного построения множеств понятийных структур, отношений между ними и формально-логического описания, в совокупности составляющих базу знаний (БЗ) предметной области (ПдО) [1]. Её объём даже для небольших ПдО может составлять несколько тысяч записей. Создание вручную такой БЗ проблематично. Эффективные и доступные системы автоматизированного проектирования неизвестны (за небольшим исключением) даже для англоязычных приложений. Для российского и украинского сегментов указанные системы просто отсутствуют. Поэтому, разработка интеллектуальной программной системы (далее – инструментального комплекса онтологического назначения (ИКОН)) автоматизированного построения онтологических баз знаний ПдО является важной и актуальной. Кроме того, полезность ИКОН существенно возрастёт, если включить в обработку украинские и российские естественно-языковые источники предметных знаний.

**Архитектура инструментального комплекса**

ИКОН автоматизированного построения онтологических баз знаний ПдО является системой, реализующей одно из направлений комплексной технологии *Data Mining*, а именно – анализ и обработку больших объёмов неструктурированных данных, в частности лингвистических корпусов текстов на украинском и/или русском языках, извлечение из последних предметных знаний с последующим их представлением в виде системно-онтологической структуры или онтологии предметной области (О ПдО) [1, 2]. Он предназначен для реализации ряда информационных технологий (ИТ), основными из которых являются:

– автоматическая обработка естественно-языковых текстов (Natural Language Processing) [3];

– извлечение из множества текстовых документов знаний, релевантных заданной ПдО, их системно-онтологическая структуризация и формально-логическое представление на одном (или нескольких) из общепринятых языков описания онтологий (Knowledge Representation).

ИКОН состоит из трёх подсистем и представляет собой интеграцию разного рода информационных ресурсов, программно-аппаратных средств обработки информации и естественного интеллекта (ЕИ), которые реализуют совокупность алгоритмов автоматизированного, итерационного построения понятийных структур предметных знаний, их накопления и/или системной интеграции. Обобщённая блок-схема ИКОН представлена на рис. 1. На нём приняты следующие сокращения: ЛКТ – лингвистический корпус текстов, ТД – текстовый документ.

*Информационный ресурс* среды включает блок формирования ЛКТ, базы данных обработки языковых структур и библиотеки понятийных структур. Первый компонент представляет собой различные источники текстовой информации, поступающей на обработку в систему. Второй компонент представляет собой различные базы данных обработки языковых структур, часть из которых формируется (наполняется данными) в процессе обработки ТД, а другая часть формируется до процесса построения О ПдО и, по сути, является электронной коллекцией различных словарей. Третий компонент представляет собой совокупность библиотек понятийных структур разного уровня представления (от наборов терминов и понятий до высокоинтегрированной онтологической структуры междисциплинарных знаний) и является результатом реализации некоторого проекта (проектирования онтологии ПдО и/или системной интеграции онтологий).

В качестве языка программирования для разработки ИКОН был выбран объектно-ориентированный язык Java [4] и свободная интегрированная среда разработки приложений (IDE) на языках программирования Java, JavaFX, Python, PHP, JavaScript, C++ и ряде других – Netbeans IDE [5].

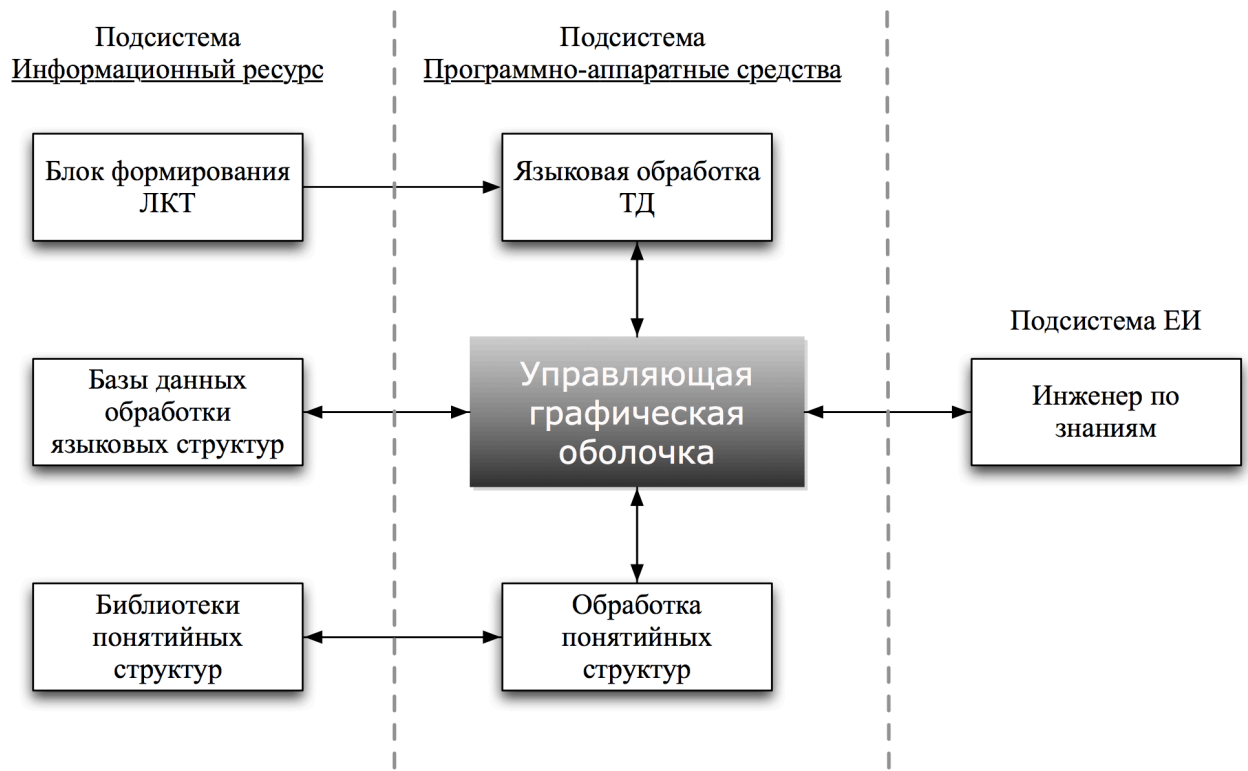

Рис.1. Обобщённая блок-схема ИнКом

*Программно-аппаратные средства* включают блоки обработки языковых и понятийных структур и *управляющую графическую оболочку (УГО)*. Последняя, во взаимодействии с естественным интеллектом, осуществляет общее управление процессом реализации связанных информационных технологий.

*Подсистема ЕИ* осуществляет подготовку и реализацию процедур предварительного этапа проектирования, в процессе основного цикла проектирования осуществляет контроль и оценку (assessment) результатов выполнения отдельных этапов проекта, принимает решение (возможно с экспертом в заданной ПдО) о степени завершённости последних и (в случае необходимости) повторении некоторых из них. Ниже описана программная реализация УГО.

**Программная реализация управляющей графической оболочки**

В состав программно-аппаратных средств ИКОН входит управляющая графическая оболочка, в которую включены функции управления всеми компонентами ИКОН и системы в целом. Структурная схема УГО ИКОН представлена на рис. 2. На нём приняты следующие сокращения: GUI – graphical user interface, OWL – Web Ontology Language, XML – Extensible Markup Language.

УГО взаимодействует с подсистемой ЕИ, осуществляет общее управление процессом реализации связанных информационных технологий.

УГО выполняет следующие функции:
– во взаимодействии с инженером по знаниям осуществляет предварительное наполнение среды материалами электронных коллекций энциклопедических, толковых словарей и тезаурусов, описывающих домен предметных знаний;
– обеспечивает запуск и последовательность исполнения прикладных программ, реализующих составные информационные технологии проектирования онтологии ПдО и системной интеграции междисциплинарных знаний (примером составной технологии является автоматизированное построение тезаурусов ПдО для поисковой системы);
– отображает ход процесса проектирования;
– содержит позиции меню для запуска, как последовательностей, так и отдельных прикладных программ, используемых в процессе проектирования;
– обеспечивает интерфейс с подсистемой ЕИ;
– индицирует сообщения о текущем состоянии проекта, его наполнении информационными ресурсами;
– обеспечивает обмен информацией между прикладными программами и базами данных через общую информационную шину.

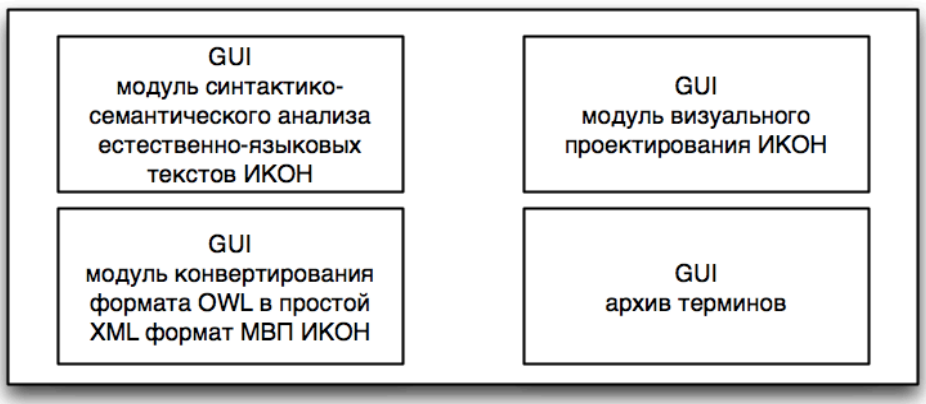

Рис. 2. Структурная схема УГО ИКОН

Одно из главных преимуществ языка программирования Java – кроссплатформенность. Таким образом, один и тот же код можно запускать под управлением операционных систем Windows, Linux, FreeBSD, Mac OS и др. Это очень важно, когда программы загружаются посредством глобальной сети Интернет и используются на различных компьютерных платформах, включая суперкомпьютерные. Язык Java содержит большое количество уже готовых классов для разработки программного обеспечения [6], благодаря чему написание кода программы требует меньшее количество временных затрат.

Для создания графического интерфейса пользователя инструментальным комплексом использована библиотека графических компонентов Swing [7, 8]. Swing относится к набору библиотек классов на языке Java – Java Foundation Classes (JFC), предоставляющих программам на Java удобный API для создания графического интерфейса пользователя. В состав JFC входят, в частности, библиотека Swing, Java 2D, AWT, Drag & Drop – API и др. Swing поддерживает специфические динамически подключаемые стили (look-and-feel) [8, 9]. Благодаря look-and-feel возможна адаптация к графическому интерфейсу платформы, на которой запускается Java-программа (т.е. к компоненту можно динамически подключить другой, специфический для операционной системы, в том числе и созданный программистом, стиль). Таким образом, приложения, использующие Swing, могут выглядеть как "родные" приложения для данной операционной системы.

Одним из основных дополнительных компонент, используемых при создании УГО ИКОН, является свободно распространяемая Java-реализация Document Object Model (DOM) для eXtensible Markup Language (XML) – библиотека Java Document Object Model (JDOM) [10], созданная с учётом особенностей языка и платформы Java. JDOM является уникальным Java-инструментом для работы с XML.

На рис. 3 представлено окно главного меню УГО, которое содержит элементы управления следующими функциями и модулями ИКОН:
– модулем конвертирования формата OWL в простой XML формат КВП (рис. 4).
– модулем визуального проектирования (КВП) (рис. 5);
– модулем синтактико-семантического анализа естественно-языковых текстов (рис. 6);

Функционирование УГО ИКОН рассмотрим на примере анализа небольшого фрагмента естественно-языкового текста из предметной области "информационные технологии".
1   Склад обчислювальної системи.
2   Характерною їх відмінністю від механічних пристроїв є те, що вони реєструють не переміщення елементів конструкції, а їх стани.
3   Це надає зручність опрацювання даних, поданих у двійковій системі числення.
4   Сукупність пристроїв, призначених для автоматизації опрацювання даних, називають обчислювальною технікою.
5   Центральним пристроєм обчислювальної системи, як правило, є комп'ютер.
6   В сучасному розумінні, комп'ютер – це універсальний електронний пристрій, призначений для автоматизації накопичення, збереження, опрацювання, передачі та відтворення даних.
7   Склад обчислювальної системи називається конфігурацією.

Для каждого предложения текста строятся деревья зависимостей, в которых определяются синтактико-семантические отношения между лексемами. Далее автоматически выделяются из текста термины предметной области в соответствии с заданными *шаблонами*, например: аббревиатура, существительное, прилагательное – существительное, существительное – прилагательное – существительное и типами синтактико-семантических отношений: объектное, принадлежность (между двумя существительными), определительное (между прилагательным и существительным), однородные слова. В качестве терминов, используемых в формальном представлении текста, автоматически выбираются преимущественно многословные словосочетания, в которых значения существительных уточняются с помощью связанных прилагательных. Приведем фрагмент *списка терминов*, состоящего из отдельных слов и словосочетаний: ПК, обчислювальна техніка, механічний пристрій, конфігурація, автоматизація, двійкова система числення, електронний пристрій, конструкція (рис. 6). Вопросы, связанные с методикой и программной поддержкой процесса построения онтологии, являются предметом отдельного рассмотрения.

Полученные в результате автоматического анализа текста множества понятий и отношений между ними являются исходными данными для построения онтологии ПдО (рис. 5).

**Заключение**

В докладе рассмотрена архитектура инструментального комплекса автоматизированного построения онтологических баз знаний предметных областей, которая, в частности, включает компоненту управляющей графической оболочки. Описана программная модель УГО, которая включает элементы управления модулями синтактико-семантического анализа естественно-языковых текстов, визуального проектирования и конвертирования формата OWL в простой XML формат КВП. В настоящее время программная реализация ИКОН, в том числе и УГО, находится в стадии завершения.

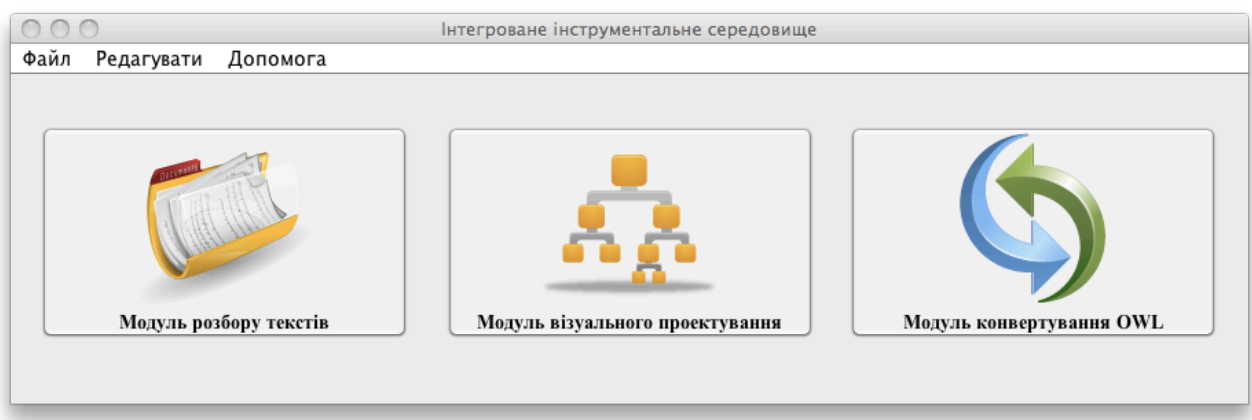

Рис. 3. Окно главного меню УГО ИКОН

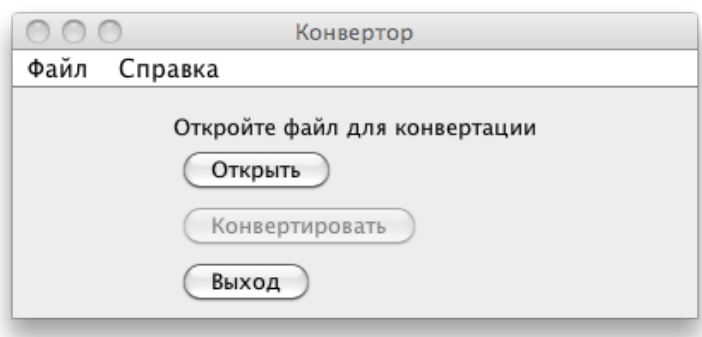

Рис. 4. Модуль конвертирования формата OWL в простой XML формат КВП ИКОН

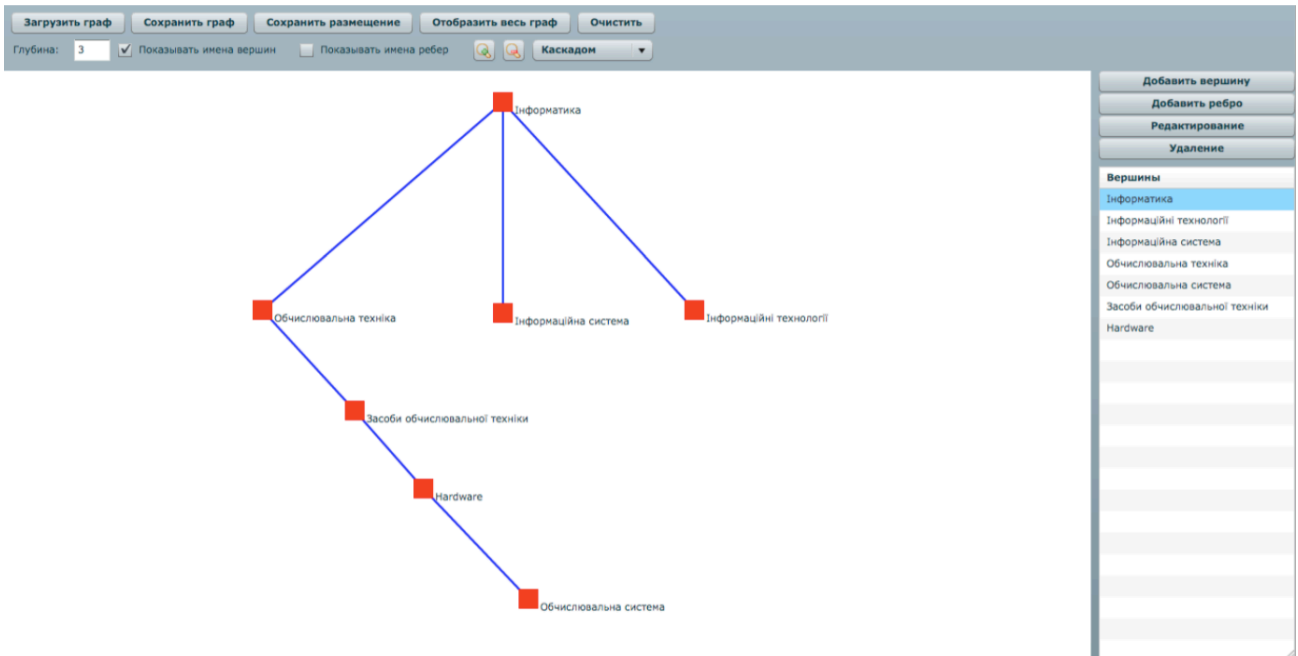

Рис. 5. Модуль визуального проектирования ИКОН

![Модуль синтактико-семантичного аналізу природно-мовних текстів]

Рис. 6. Модуль синтактико-семантического анализа естественно-языковых текстов

**Палагін О.В., Петренко М.Г., Величко В.Ю., Малахов К.С. До питання розробки онтолого-керованої інтелектуальної програмної системи.**

В роботі описана архітектура інтелектуальної програмної системи автоматизованої побудови онтологічних баз знань предметних областей та програма модель підсистеми керуючої графічної оболонки.

Ключові слова: онтологія предметної області, інструментальний комплекс онтологічного призначення, керуюча графічна оболонка.

**Palagin A.V., Petrenko N.G., Velichko V.U., Malakhov K.S. The problem of the development ontology-driven architecture of intellectual software systems.**

The paper describes the architecture of intelligence system for automated construction of ontological knowledge bases of subject areas and the programming model subsystem management GUI.

Keywords: domain ontology, tool complex ontological destination, GUI management.